\documentclass[]{gmaart2019}

\renewcommand{\varProcName}{Proc. 31. Workshop Computational Intelligence}
\renewcommand{\varLocation}{Berlin}
\renewcommand{\varDate}{25.-26.11.2021}

\usepackage[utf8]{inputenc} 
\usepackage[T1]{fontenc} 
 
\usepackage{amsmath,amssymb,amsfonts,mathtools,amsthm} 

\usepackage[babel,autostyle]{csquotes}

\usepackage{tabularx}
\newcolumntype{L}[1]{>{\raggedright\arraybackslash}p{#1}} 
\newcolumntype{C}[1]{>{\centering\arraybackslash}p{#1}} 
\newcolumntype{R}[1]{>{\raggedleft\arraybackslash}p{#1}} 
\usepackage{booktabs}

\usepackage{paralist}   

\usepackage[noend]{algpseudocode}

\algrenewcommand\algorithmicrequire{\textbf{Voraussetzung:}}
\algrenewcommand\algorithmicensure{\textbf{Abschlussbedingung:}}

\usepackage{listings} 
\usepackage{subcaption}  

\usepackage{epstopdf}
\usepackage{xcolor}
\selectcolormodel{gray}  

\usepackage{scrhack}
\usepackage{varwidth}
\usepackage{rotating} 
\usepackage[defaultlines=2,all]{nowidow}  
\usepackage{import}
\usepackage{placeins}


\usepackage[acronym]{glossaries}
\usepackage{xspace}
\usepackage{xcolor,colortbl}
\usepackage{moresize}
\usepackage{hyperref}
\usepackage{cleveref}
\usepackage{multirow}

\usepackage{etoolbox}
\preto\section\glsresetall

\newacronym{DNN}{DNN}{Deep Neural Network}
\newacronym{FCN}{FCN}{Fully Connected Network}
\newacronym{CNN}{CNN}{Convolutional Neural Network}
\newacronym{MAE}{MAE}{Mean Absolute Error}

\begin{document}


\hyphenpenalty=2000

\pagenumbering{roman}
\setcounter{page}{1}
\pagestyle{scrheadings}
\pagenumbering{arabic}



\setnowidow[2]
\setnoclub[2]

\newcommand\stress[1]{{\color{red}#1}}
\newcommand\ie{i.\,e.\xspace}
\newcommand\eg{e.\,g.\xspace}
\newcommand\etal{et\,al.\xspace}

\DeclareRobustCommand{\bbone}{\text{\usefont{U}{bbold}{m}{n}1}}
\newcommand{\CC}[1]{\cellcolor{gray!#1}}
\definecolor{Gray}{gray}{0.85}
\definecolor{LightCyan}{rgb}{0.88,1,1}

\renewcommand{\Title}{Smart Data Representations: \\ Impact on the Accuracy of \\ Deep Neural Networks}

\renewcommand{\Authors}{
    Oliver Neumann\textsuperscript{1}, Nicole Ludwig\textsuperscript{2}, Marian Turowski\textsuperscript{1}, \\
    Benedikt Heidrich\textsuperscript{1}, Veit Hagenmeyer\textsuperscript{1}, Ralf Mikut\textsuperscript{1}
}
\renewcommand{\Affiliations}{
    \textsuperscript{1} Institute for Automation and Applied Informatics, \\ Karlsruhe Institute of Technology \\
    Hermann-von-Helmholtz-Platz 1, 76344 Eggenstein-Leopoldshafen \\
    \textsuperscript{2} Cluster of Excellence Machine Learning, \\ University of Tübingen \\
    Maria-von-Linden Str. 6, 72076 Tübingen \\
    E-Mail: oliver.neumann@kit.edu \\
}

\renewcommand{\AuthorsTOC}{O. Neumann, N. Ludwig et al.} 
\renewcommand{\AffiliationsTOC}{Karlsruhe Institute of Technology, Institute for Automation and Applied Informatics; Cluster of Excellence Machine Learning, University of Tübingen} 

\setLanguageEnglish
							 
\setupPaper 

\section*{Abstract}

\Glspl{DNN} are able to solve many complex tasks with less engineering effort and better performance. However, these networks often use data for training and evaluation without investigating its representation, i.e.~the form of the used data. In the present paper, we analyze the impact of data representations on the performance of \Glspl{DNN} using energy time series forecasting.
Based on an overview of exemplary data representations, we select four exemplary data representations and evaluate them using two different \Gls{DNN} architectures and three forecasting horizons on real-world energy time series. The results show that, depending on the forecast horizon, the same data representations can have a positive or negative impact on the accuracy of \Glspl{DNN}.

\section{Introduction}\label{sec:Introduction}

\Glspl{DNN} can better solve complex tasks such as image classification \cite{liu_scene_2018, kamble_handwritten_2015}, object detection \cite{zhu_orientation_2015}, or instance segmentation \cite{jader_deep_2018, scherr_cell_2020} with less effort than traditional approaches. Nevertheless, \glspl{DNN} require data for training and evaluation. However, data is often used by \glspl{DNN} without further investigation of different data representations.

Since data representations influence what \glspl{DNN} learn and which architectures can be used, data representations should be investigated further. The representation of the data can be changed through transformations such as reshaping, aggregation, or selection. Although recent literature, including work on feature engineering \cite{faust_intelligent_2019, seide_feature_2011, zimbra_brand-related_2016}, introduces new data representations and compares them \cite{monti_geometric_2017, zhou_graph_2020, jahankhani_eeg_2006, jia_solving_1995, heidrich_forecasting_2020}, it does not systematically investigate the influence of data representations on the performance of \glspl{DNN}.

In this paper, we analyze the impact of data representations on the performance of \glspl{DNN} at the commonly investigated example of energy time series forecasting (see e.g.~\cite{ordiano_energy_2018, imani_sequence_2019, hafeez_electric_2020, heidrich_forecasting_2020}). For this purpose, we investigate the time series in its original form and the derivative of the time series, and both reshaped as an image. For the analysis, we use two different architectures, namely a \Gls{FCN} and a Convolutional Neural Network (CNN).

The remainder of the paper is structured as follows. In the second chapter, we introduce different transformations for data representations. In the third chapter, we present an energy forecasting use case to demonstrate the impact of different data representations. In the fourth chapter, we discuss the findings of this paper, before we finally give a conclusion.

\section{Transformations for Data Representations}\label{sec:DataRepresentations}

\begin{figure}[!ht]
	\centering
	\includegraphics[width=0.75\linewidth]{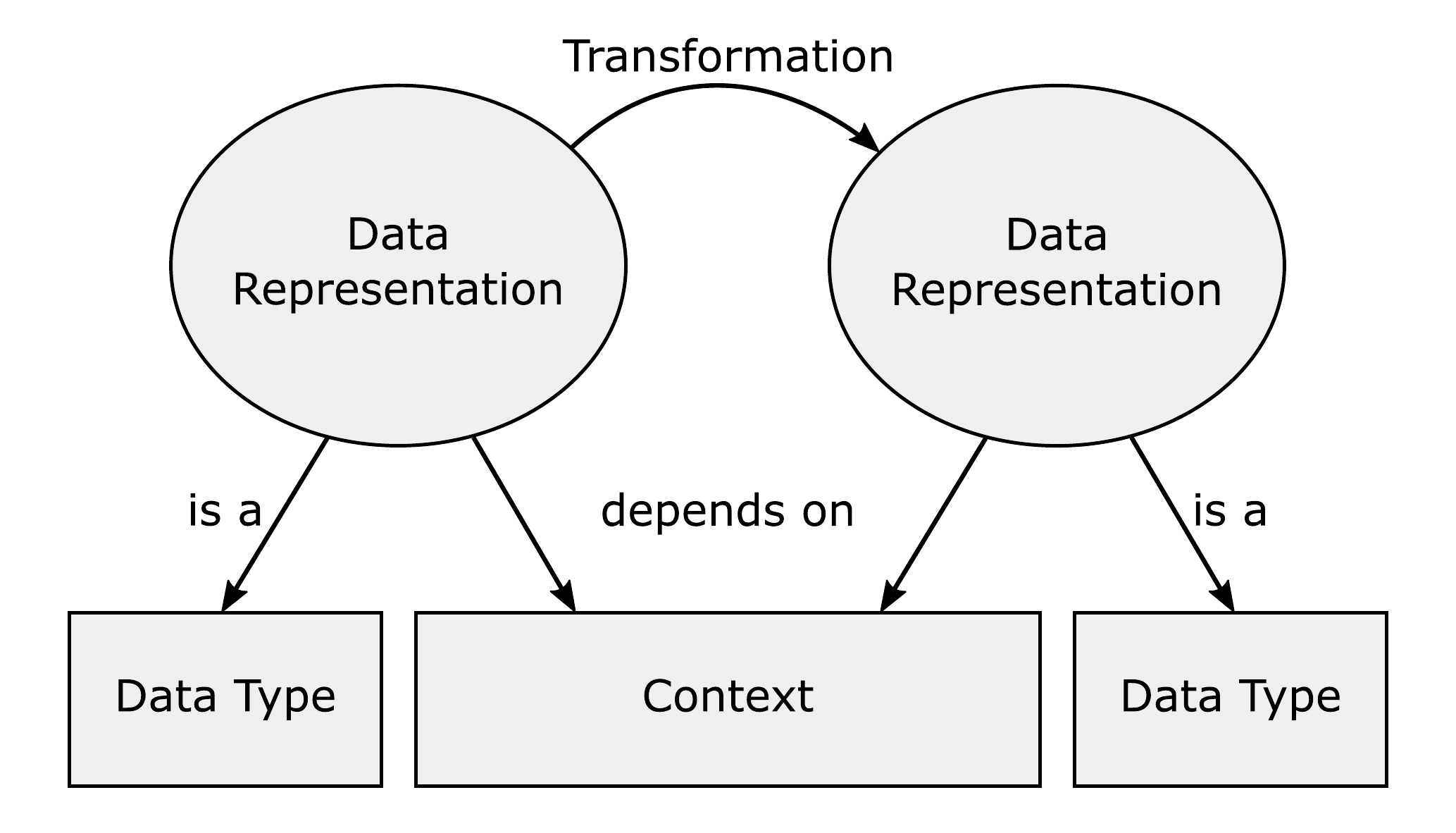}
	\caption{Relationship between data representation, data type, transformation, and context. A transformation converts a data representation into another, while a data representation is a data type and depends on a context.}
	\label{fig:overview}
\end{figure}

In this chapter, we introduce different transformations for changing data representations based on the data types vectors or matrices. Transformations allow to convert data from one data representation to another. Unlike data types, data representations are context dependent and thus are typically difficult to characterize. For comprehensibility, we thus present transformations, which are per se context independent (see \Cref{fig:overview}).
For the present paper, we consider reshaping, selection, aggregation, differences, convolution, rescaling, clustering, and latent space transformation as exemplary transformations for converting data from one representation into another. For each transformation, we briefly describe its key idea and underlying concept.

\paragraph{Reshaping} Vectors and matrices can be transformed by changing their shape. This reshaping allows the use of different model architectures. Reshaping is defined by a function that maps each element of a vector or matrix to a resulting vector or matrix with a different dimensionality.

\paragraph{Selection} Data representations can be transformed by selecting specific elements. To select certain elements, one can specify the related indices, which then define a subset of the considered vector or matrix. Choosing specific elements of a vector or matrix can be beneficial because a data representation can contain unnecessary information.

\paragraph{Aggregation} The aggregation of data can be used as a transformation for data representations. Aggregations can help a DNN to achieve higher performances because aggregations can, for example, reduce the dimensionality and noise in the data. They can be applied on single vectors or matrices along one or more axes, leading to a matrix or vector depending on the dimensionality of the input.

\paragraph{Differences} Calculating the differences is similar to the discrete derivation and can be applied to vectors and matrices. The data is transformed by subtracting the values of certain points or axes, vectors, or matrices depending on the input data representation. For vector inputs, the difference between certain points is calculated. For matrix inputs, the differences are calculated for certain subvectors or submatrices depending on the dimensionality of the matrix and along which axis the differences should be calculated.

\paragraph{Convolution} A convolution is a mathematical operation that combines two functions. The convolution, e.g.~a frequency filter, can be described by a kernel that is multiplied iteratively over the input data and summed afterward. Both the kernel and the input data can be vectors or matrices.

\paragraph{Rescaling} The representation of data can be transformed by fitting a function on the data and resampling from that function. Thereby, data can be upscaled or downscaled, where upscaling increases and downscaling decreases the amount of data. Exemplary methods to approximate the underlying function are linear, cubic, or spline interpolation.

\paragraph{Clustering} Data can also be clustered such that it is represented by cluster representatives. The cluster representatives are determined by similarity measures based on, for example, density or distances. Common clustering approaches are $k$-Means \cite{chinrungrueng_optimal_1995}, fuzzy $c$-means \cite{bezdek_fcm_1984}, BIRCH \cite{zhang_birch_1996}, OPTICS \cite{ankerst_optics_1999}, or DBSCAN \cite{ester_density-based_1996}.

\paragraph{Latent Space Transformation} Latent space transformations learn a latent data representation of a dataset. This latent data representation has a lower dimensionality as the original dataset and could be a vector or matrix data representation. There are several approaches to use the latent space information and reduce the dimensionality like Principal Component Analysis \cite{pearson_lines_1901}, Linear Discriminant Analysis \cite{cohen_applied_2014}, or Autoencoder \cite{kramer_nonlinear_1991}.

\section{Energy Forecasting Use Case}\label{sec:UseCase}

In this chapter, we show how data representations affect the forecasting accuracy of a \gls{DNN} when forecasting the German electricity demand and using four different data representations, namely \emph{naive}, \emph{naive differences}, \emph{reshaped}, and \emph{reshaped differences}. In the following, we first describe the data for this use case and the data representations. Afterward, we present the selected baselines and DNN architectures to forecast the electricity demand. In the last section, we present the results and compare the evaluated data representations.

\subsection{Data}\label{sec:Data}

For the electricity demand, we use data from the European Network of Transmission System Operators for Electricity provided by Open Power System Data \cite{wiese_open_2019}. We select the electricity demand for Germany from the beginning of 2015 up to the end of 2019, which is the last complete year of data. The data contain typical daily, weekly, and seasonal patterns, which we account for with the help of calendar information. As calendar information, we use hour, day of week, day of year, weekend, and holiday, where the first three are encoded as sine and cosine functions and the last two are encoded as Boolean variables.

\subsection{Evaluated Data Representations}\label{sec:EvaluatedDataRepresentations}

This section describes the four data representations chosen to analyze the impact of data representations on the forecasting performance of \Glspl{DNN}, i.e.~the \textit{naive}, the \textit{naive differences}, the \textit{reshaped}, and the \textit{reshaped differences} data representations. Overall, this selection results in two vector-based and two matrix-based data representations. These data representations are used as input for the evaluated \Gls{DNN} architectures, whose output also depend on these representations. 

\begin{figure}[!ht]
	\centering
	\includegraphics[width=\linewidth]{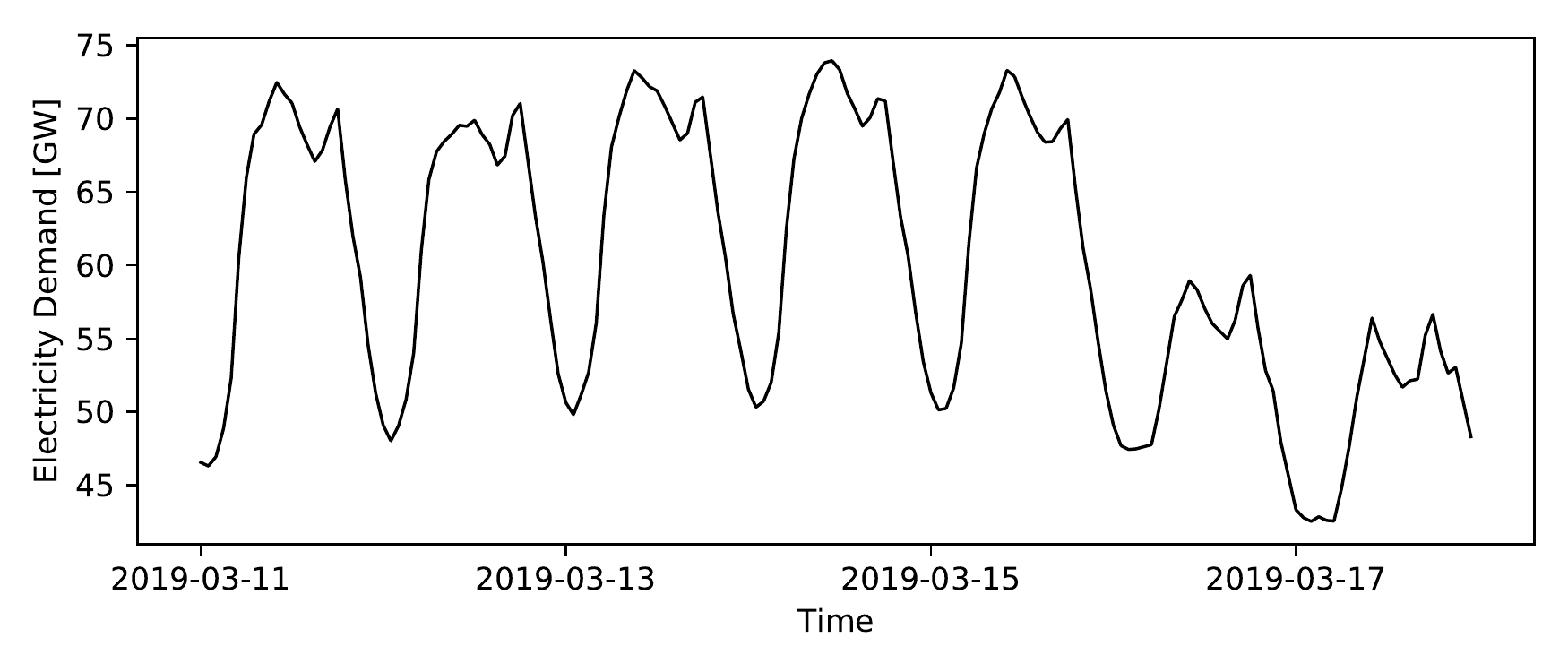}
	\caption{The \textit{naive} data representation that comprises the electricity demand of the past 168 hours shown for the exemplary forecast origin 18.03.2019.}
	\label{fig:naive_example}
\end{figure}

\paragraph{Naive}
The \textit{naive} data representation comprises the vector of the last 168 hours of the electricity demand (see \Cref{fig:naive_example}). It is defined as

\begin{equation}
    \begin{bmatrix}
    x_k \\
    \vdots \\
    x_{k - 167}
    \end{bmatrix},
\end{equation}

where $x \in X$ is the set of historical electricity demand values and $k$ the forecast origin.

\begin{figure}[!ht]
	\centering
	\includegraphics[width=\linewidth]{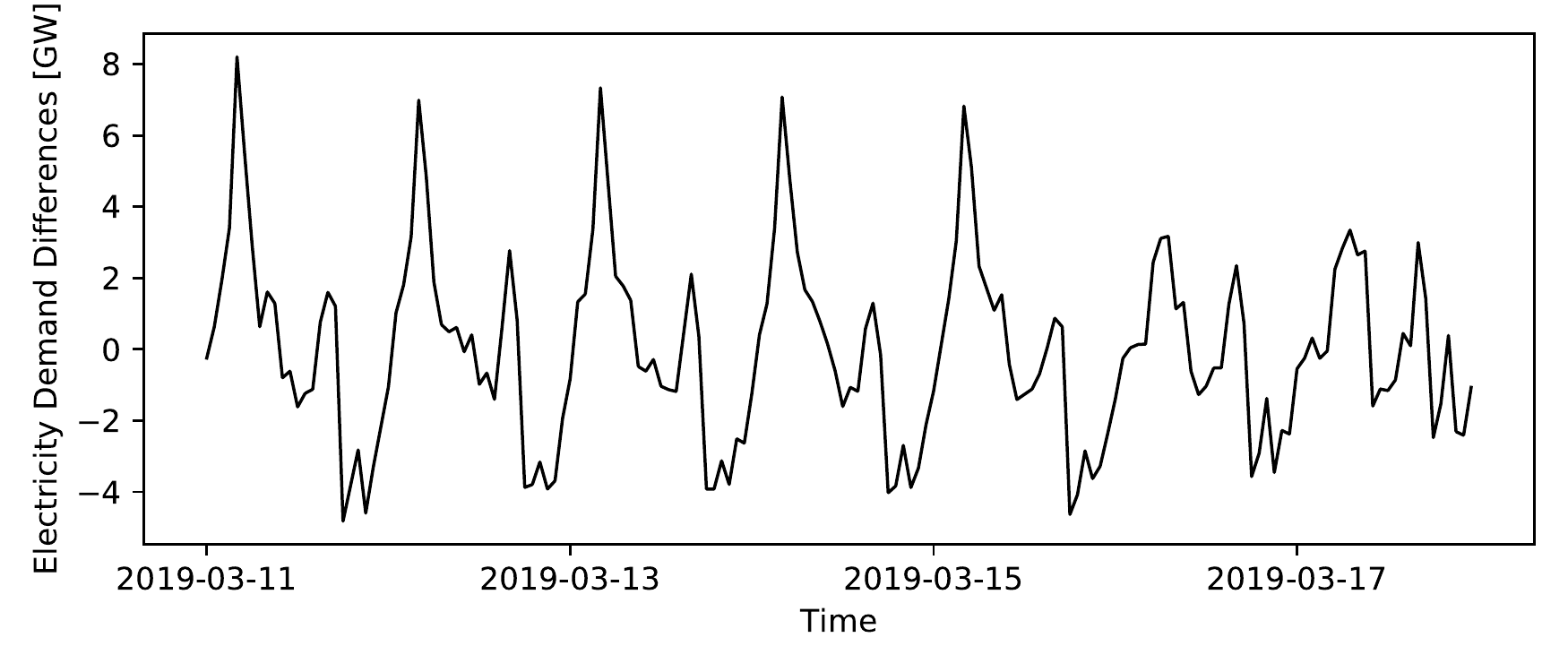}
	\caption{The \textit{naive differences} data representation that comprises the electricity demand differences with $h=1$ of the past 168 hours for the exemplary forecast origin 18.03.2019.}
	\label{fig:naive_differences_example}
\end{figure}

\paragraph{Naive Differences}
The \textit{naive differences} data representation is again comprised of a vector of the last 168 hours of the electricity demand. However, instead of using the raw values as in the \textit{naive} data representation, we now use differences. The lag of these differences depends, in our case, on the forecast horizon, e.g., for a one-day ahead forecast, we calculate differences with a lag of one day. An exemplary week is illustrated in \Cref{fig:naive_differences_example} and the data representation is then defined as

\begin{equation}
    \begin{bmatrix}
    x_{k} - x_{k - h} \\
    \vdots \\
    x_{k - 167} - x_{k - 167 - h}
    \end{bmatrix},
\end{equation}

where $x \in X$ is the set of historical electricity demand values, $k$ the forecast origin, and $h$ the lag used for differencing that we set as the forecast horizon (e.g.~1, 24, 168).

\begin{figure}[!ht]
	\centering
	\includegraphics[width=\linewidth]{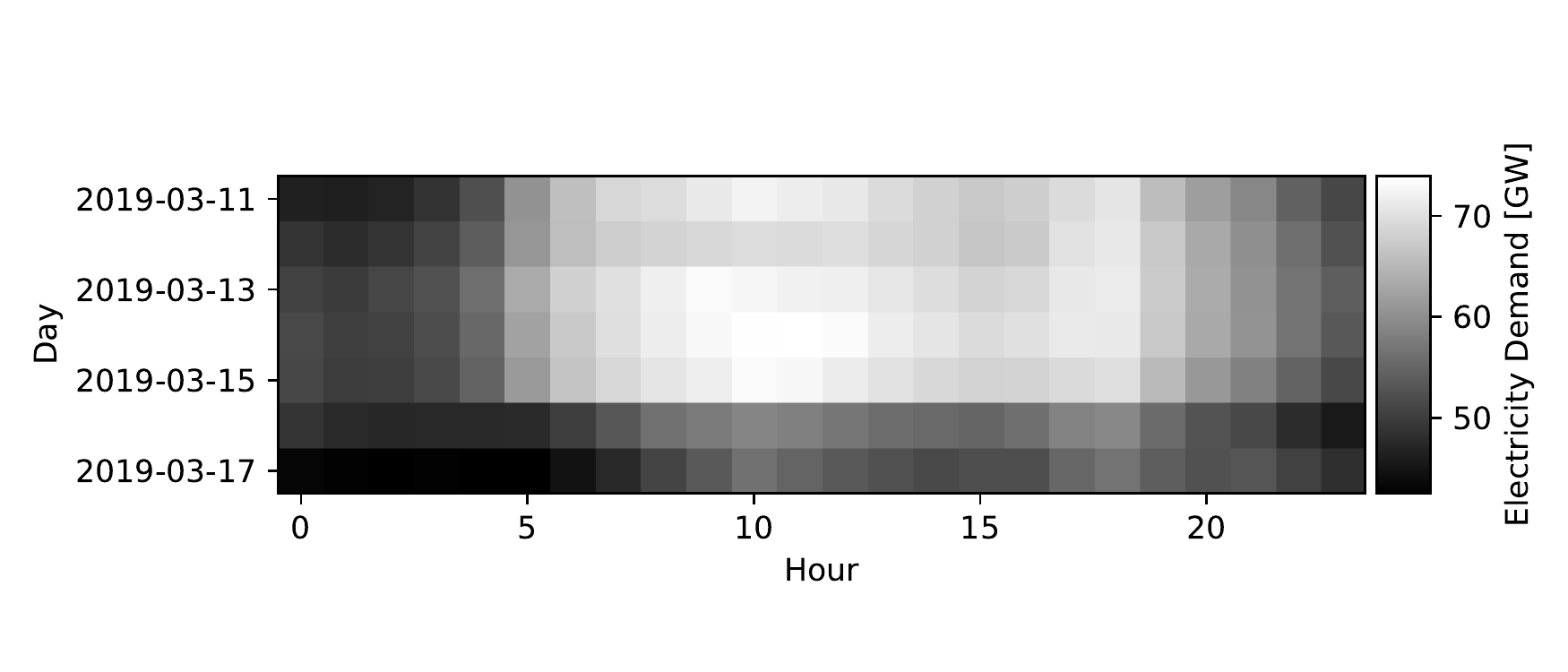}
	\caption{The \textit{reshaped} data representation that comprises the electricity demand of the past 168 hours shown for the exemplary forecast origin 18.03.2019.}
	\label{fig:reshaped_example}
\end{figure}

\paragraph{Reshaped}
The \textit{reshaped} data representation uses the \textit{naive} data representation and reshapes it into a two-dimensional matrix such that each row represents a day. Consequently, the \textit{reshaped} data representation consists of a 7x24 dimensional matrix (see \Cref{fig:reshaped_example}) and is defined as

\begin{equation}
    \begin{bmatrix}
    x_k \\
    \vdots \\
    x_{k - 167}
    \end{bmatrix}
    \Leftrightarrow
    \begin{bmatrix}
    x_{k} & \hdots &  x_{k - 23} \\
    \vdots & \ddots & \vdots \\
    x_{k - 144} & \hdots &  x_{k - 167}
    \end{bmatrix},
\end{equation}

where $x \in X$ is the set of historical electricity demand values and $k$ the forecast origin.

\begin{figure}[!ht]
	\centering
	\includegraphics[width=\linewidth]{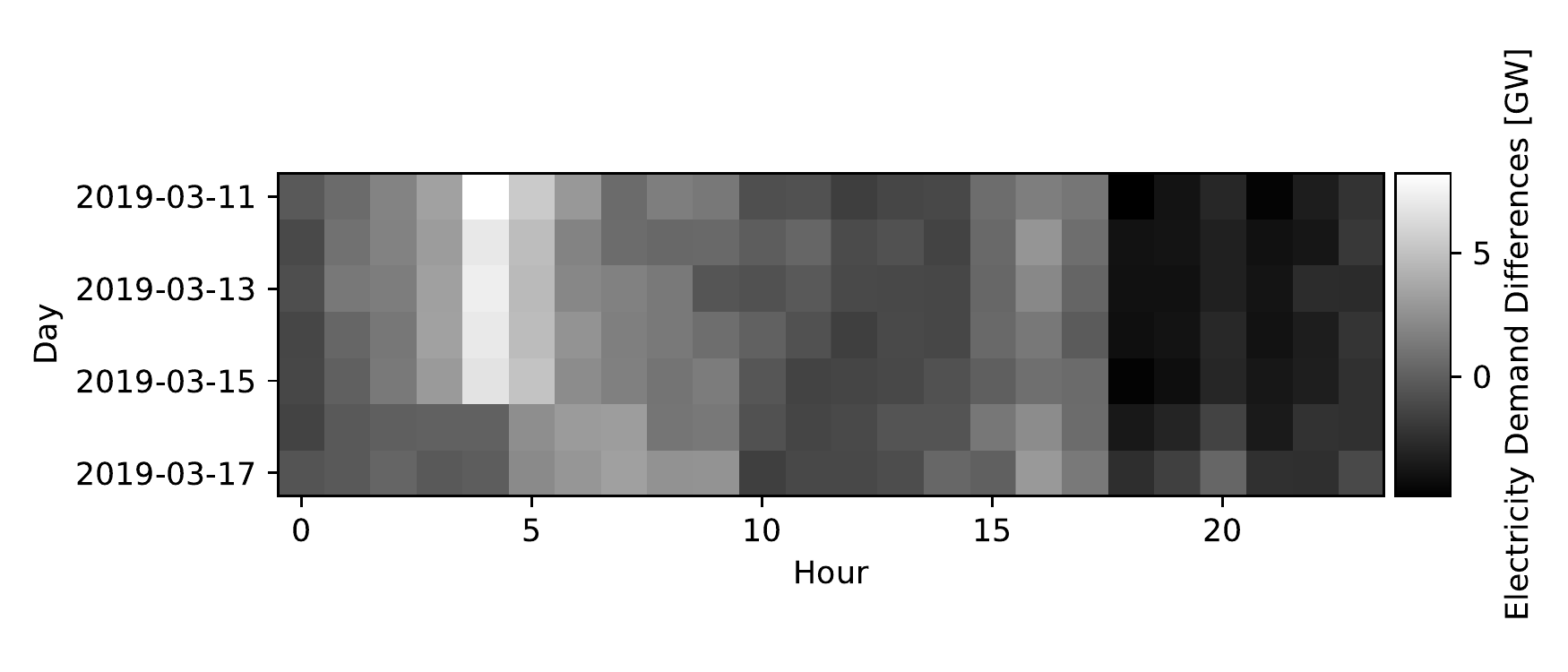}
	\caption{Exemplary week of the \textit{reshaped differences} data representation that reshapes the electricity demand differences with $h=1$ of the past 168 hours as a 24x7 matrix for the exemplary forecast origin 18.03.2019.}
	\label{fig:reshaped_differences_example}
\end{figure}

\paragraph{Reshaped Differences}
The \textit{reshaped differences} data representation is equivalent to the \textit{reshaped} data representation but uses the \textit{naive differences} data representation as its basis. The \textit{reshaped differences} data representation, therefore, also consists of a 7x24 dimensional matrix (see \Cref{fig:reshaped_differences_example} for an exemplary week) and is defined as

\begin{equation}
    \begin{bmatrix}
    x_{k} - x_{k - h} \\
    \vdots \\
    x_{k - 167} - x_{k - 167 - h}
    \end{bmatrix}
    \Leftrightarrow
    \begin{bmatrix}
    x_{k} - x_{k - h} & \hdots &  x_{k - 23} - x_{k - 23 - h} \\
    \vdots & \ddots & \vdots \\
    x_{k - 144} - x_{k - 144 - h} & \hdots &  x_{k - 167} - x_{k - 167 - h}
    \end{bmatrix},
\end{equation}

where $x \in X$ is the set of historical electricity demand values, $k$ the forecast origin, and $h$ the lag used for differencing that we set as the forecast horizon (e.g.~1, 24, 168).

\subsection{Experimental Setup}\label{sec:setup}
In this section, we introduce the evaluated \Gls{DNN} architectures and the selected baselines before we describe the experimental setup including the train-validation-test split, the considered forecast horizons, the number of performed runs, the used evaluation metrics, and the implementation.

\begin{figure}[!ht]
	\centering
    \begin{subfigure}[b]{0.47\textwidth}
		\centering
		\includegraphics[width=\linewidth]{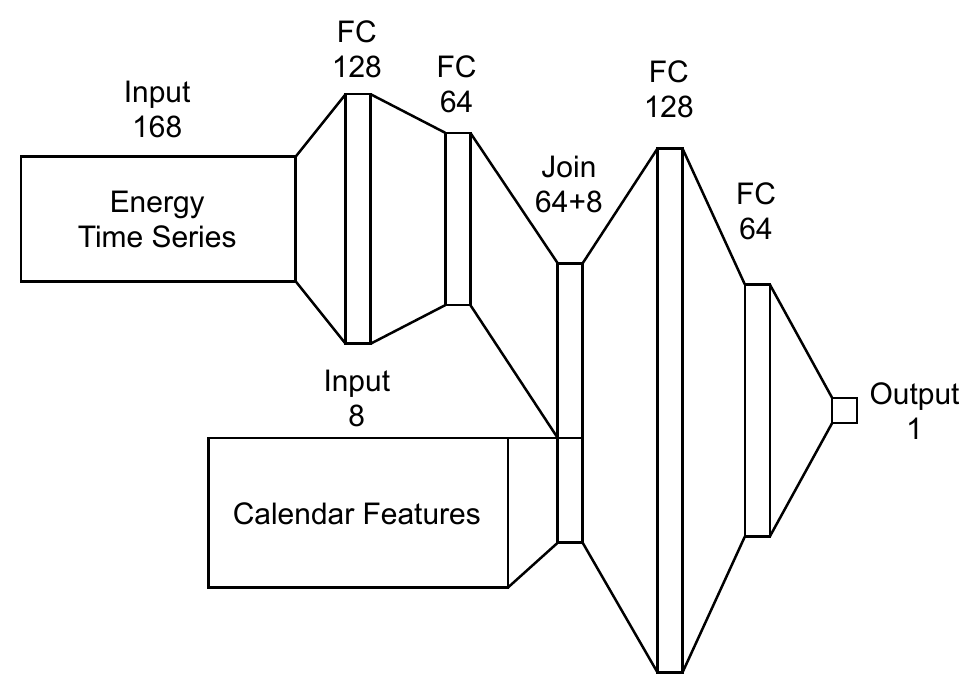}
		\caption{Fully Connected Network (FCN)}
		\label{fig:fc_arch}
	\end{subfigure}
	\begin{subfigure}[b]{\textwidth}
		\centering
		\includegraphics[width=\linewidth]{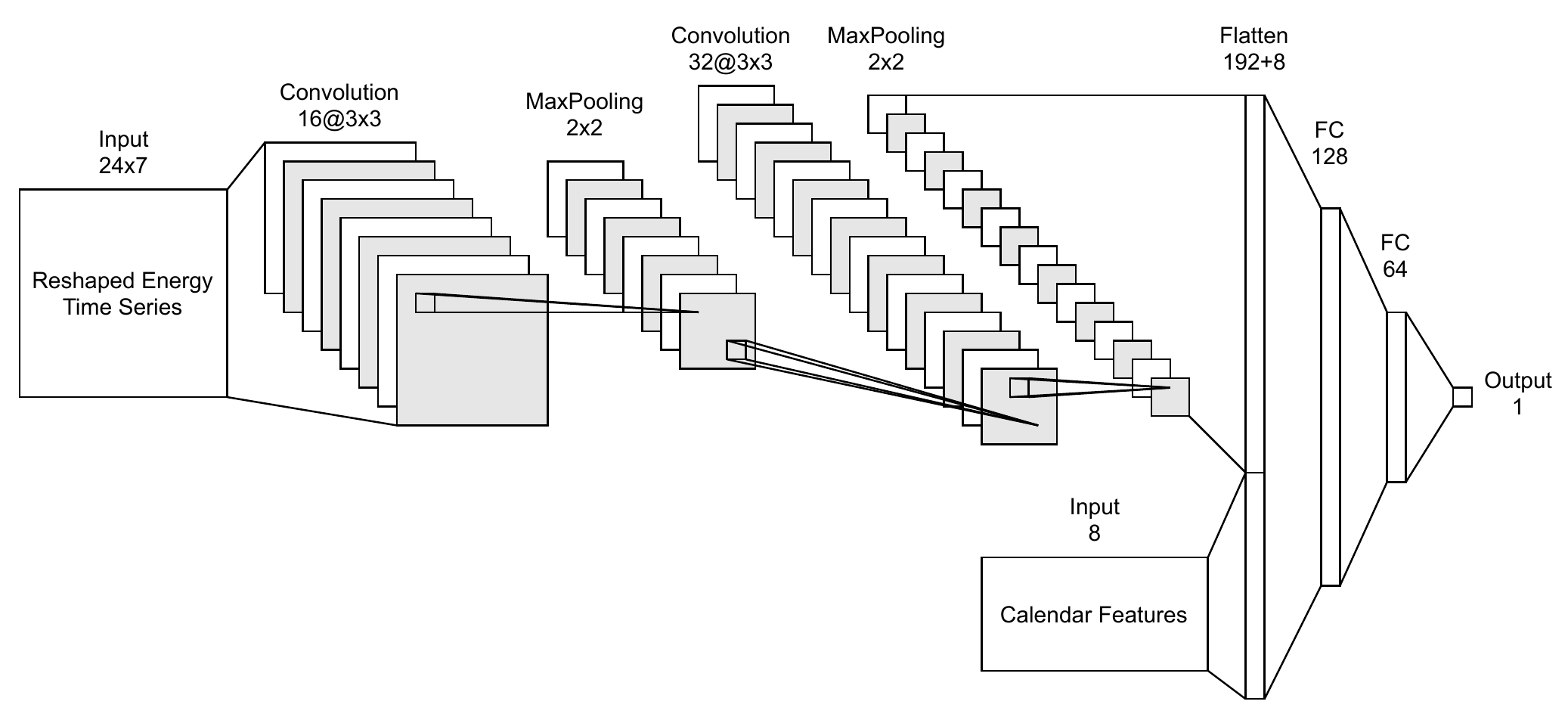}
		\caption{Convolutional Neural Network (CNN)}
		\label{fig:cnn_arch}
	\end{subfigure}
	\caption{DNN architectures used for forecasting the energy time series, where the numbers indicate the number of neurons in the fully connected layers and the number of features with the corresponding kernel sizes in the convolutional layers (separated with @).}
	\label{fig:architecture}
\end{figure}

To investigate the impact of data representations in energy time series forecasting, we use two \Glspl{DNN} (see \Cref{fig:architecture}). The first \Gls{DNN} is a \Gls{FCN}. It only consists of fully connected layers, whose input is a vector of historical energy data. This \Gls{FCN} is applied to both naive data representations. The second \Gls{DNN} is a \Gls{CNN}. It consists of convolutional layers followed by fully connected layers that process the energy time series as a matrix. This CNN is applied to both reshaped data representations.

The \Gls{FCN} consists of two parts. The first part processes the energy time series vector input into a 64 dimensional latent vector representation and consists of two layers. The second part joins the 64 dimensional latent energy vector with the calendar feature vector and processes the concluding vector to the single forecast output.

The \Gls{CNN} is also split into two parts. The first part processes the reshaped energy time series with two stacked convolutional layers. This first part results in a latent representation of matrices, which is then flattened and joined with the calendar features. The second part processes the vector-based latent representation of the energy time series and calendar features with three hidden layers.

As baselines for the general forecasting result, we choose two linear regression models that forecasts the electricity demand based on selected past electricity demand or rather electricity demand difference values and calendar features. For the electricity demand, we use the same data as for the \textit{naive} data representation. Analog for the electricity demand differences, we use the same data as for the \textit{naive differences} data representation. Regarding the calendar features, we use the same calendar features as for the \Gls{FCN} and \Gls{CNN} models. Thus, the linear regression for the electricity demand is defined as

\begin{equation}
     \hat{x}_{k+h} = \alpha + \sum_{i=0}^{167} \beta_i x_{k-i} + \sum_{j=1}^{8} \gamma_j c_{k+h,j}
    \text{,}
\end{equation}

where $x \in X$ is the electricity demand, $c \in C$ the calendar features, and $h$ the forecast horizon. For the electricity demand differences the linear regression is defined as 

\begin{equation}
    \hat{\Delta}({x}_{k+h}, {x}_{k}) = \alpha + \sum_{i=0}^{167} \beta_i (x_{k-i} - x_{k-i-h}) + \sum_{j=1}^{8} \gamma_j c_{k+h,j}
    \text{,}
\end{equation}

where $x \in X$ is the electricity demand, $c \in C$ the calendar features, and $h$ the forecast horizon.

To apply the mentioned architectures and benchmarks, we run the following setup: Regarding the train-validation-test split, we use the years 2015 to 2017 for training, 2018 for validation, and 2019 for testing. With regard to the forecast horizon, we forecast a specific hour for each model, i.e.~one-hour, one-day, and one-week ahead. We evaluate the forecast horizons with the \Gls{MAE} defined by $\frac{\sum_{i=1}^{N} | y_i - \hat{y}_i |}{N}$, where $y_i$ is the ground truth and $\hat{y}_i$ the prediction. For each combination of the four evaluated data representations and the three forecast horizons, we run the respective network with ten different seeds. We report the mean and standard deviation of these ten runs and use this mean to calculate the relative advantage in percent compared to the \textit{naive} data representation, i.e.~ $\frac{MAE_{compare}}{MAE_{naive}} - 1$. 

The whole experimental setup is implemented in Python. For this purpose, we use PyTorch \cite{paszke_pytorch_2019} for realizing the \Gls{DNN} architectures and pyWATTS \cite{heidrich_pywatts_2021} for defining a reproducible and reusable pipeline. The implementation is available on GitHub\footnote{\url{https://github.com/KIT-IAI/SmartDataRepresentations}}.

\subsection{Results}\label{sec:results}

In this section, we present the results of the evaluated data representations regarding the energy time series forecast. For the 2019 test data, we report the results for the one-hour, the one-day, and the one-week ahead forecast (see \Cref{tbl:results2019}). In addition, for all forecast horizons, we consider the \textit{naive} data representation as a benchmark. More specifically, we compare the mean of all runs of each data representation to this benchmark before comparing the best data representations to the baseline.

\begin{table}[!ht]
	\centering
	\caption{Forecasting \Gls{MAE} results of the four evaluated data representations in GW for the three selected forecast horizons. For the \textit{naive} and \textit{naive differences} data representations, we use the \Gls{FCN}, while we apply the \Gls{CNN} for the \textit{reshaped} and \textit{reshaped differences} data representations.}
    \resizebox{\textwidth}{!}{
    	\begin{tabular}{ccccc}
    		\toprule
    		\multirow{2}{*}{Forecast Horizon} & \multicolumn{4}{c}{Data Representations} \\
    		 & Naive & Naive Differences & Reshaped & Reshaped Differences \\
    		\midrule
            One-Hour & 0.420 $\pm$ 0.025 & \textbf{0.376 (-10.3\%) $\pm$ 0.027} & 0.531 (+26.5\%) $\pm$ 0.015 & 0.385 (-8.3\%) $\pm$ 0.003 \\
            One-Day & \textbf{1.197 $\pm$ 0.128} & 1.289 (+7.6\%) $\pm$ 0.118 & 1.209 (+0.9\%) $\pm$ 0.017 & 1.240 (+3.5\%) $\pm$ 0.010 \\
            One-Week & \textbf{1.677 $\pm$ 0.084} & 1.775 (+5.8\%) $\pm$ 0.102 & 1.742 (+3.9\%) $\pm$ 0.028 & 1.820 (+8.5\%) $\pm$ 0.012 \\
    		\bottomrule
    	\end{tabular}
	}
	\label{tbl:results2019}
\end{table} 

For the one-hour ahead forecast, the \textit{naive differences} data representation is best with an improvement of 10\%. In contrast to this data representation, the \textit{reshaped} data representation reduces the forecasting accuracy up to 27\%. The \textit{reshaped differences} data representation performs similarly as the \textit{naive differences} data representation with an improvement of 8\% compared to the \textit{naive} data representation.

For the one-day ahead forecast, the \textit{naive} data representation performs best, while the \textit{reshaped} data representation performs quite similar with a higher \Gls{MAE} of 1\%. The \textit{naive differences}, and \textit{reshaped differences} data representations reduce the forecasting performance between 3\% to 8\%.

For the one-week ahead forecast, the \textit{naive} data representation is the best performing data representation. The \textit{reshaped} data representation has a 4\% higher \Gls{MAE}. However, both data representations based on differences, namely the \textit{naive differences} and the \textit{reshaped differences} data representations, perform worse than the \textit{naive} data representation with higher \Glspl{MAE} between 5\% and 9\%.

\begin{table}[!ht]
	\centering
	\caption{Forecasting \Gls{MAE} results of the baseline and the best evaluated data representations in GW for the three selected forecast horizons. For the baseline, we employ a linear regression. For the \textit{naive} data representation, we use the \Gls{FCN}, while we apply the \Gls{CNN} for the \textit{reshaped differences} data representation.}
    \resizebox{0.9\textwidth}{!}{
    	\begin{tabular}{ccccl}
    		\toprule
    		Forecast Horizon &
    		\multicolumn{2}{c}{Baselines} &
    		\multicolumn{2}{c}{Best Data Representation} \\
    		 & \textit{Naive} & \textit{Naive Differences} & & \\
    		\midrule
            One-Hour & 0.506 & \textbf{0.470} & 0.376 (-20.0\%) $\pm$ 0.027 & \textit{Reshaped Differences} \\
            One-Day  & \textbf{1.639} & 1.691 & 1.197 (-27.0\%) $\pm$ 0.128 & \textit{Naive} \\
            One-Week & \textbf{1.975} & 2.128 & 1.677 (-15.1\%) $\pm$ 0.084 & \textit{Naive} \\
    		\bottomrule
    	\end{tabular}
	}
	\label{tbl:benchmark}
\end{table}

For all forecast horizons, the best evaluated data representation performs better than the selected baselines. For example, the \textit{reshaped differences} data representation improves the forecasting accuracy by at least 20\% for the one-hour ahead forecast. The \textit{naive} data representation obtains a 27\% better forecasting accuracy for the one-day ahead forecast compared to the best performing baseline. For the one-week ahead forecast, the \textit{naive} data representation achieves an 15\% better forecasting accuracy. 
We additionally run the \textit{naive} and \textit{naive differences} data representation experiments on a simple Multilayer Perceptron (MLP) with one hidden layer consisting of ten neurons and achieve similar results as for the linear regression model.

\section{Discussion} \label{sec:Discussion}

This section discusses the results of the energy forecasting use case. Our results show that the evaluated data representations, despite essentially containing the same information, result in different accuracies in the energy forecasting use case. Although the \textit{reshaped} and the \textit{reshaped differences} data representations are based on a similar concept, only the \textit{reshaped differences} data representation outperforms the naive benchmark in the single case of the one-hour ahead forecast.
Furthermore, depending on the forecast horizon, the data representations perform differently. Nevertheless, there is no data representation that offers the best forecasting accuracy for all evaluated forecast horizons. For example, the \textit{naive differences} data representation is the best data representation for one-hour ahead forecasts. However, for the one-day and one-week ahead forecast, the \textit{naive} data representation performs best. As a consequence, it should be investigated in which way various data representations influence the forecasting accuracy of \Glspl{DNN} and how they are affected by different forecast horizons and architectures given a specific use case.

In the evaluated energy forecasting use case, we consider four data representations and two \gls{DNN} architectures. For these data representations and architectures, our results show that the forecasting accuracy varies. However, the ambiguous results do not allow for a general statement regarding the impact of data representations on the performance of \glspl{DNN}. Moreover, the considered use case does not provide insights on the transferability of the results to other not yet evaluated data representations and to use cases from other domains. In addition, this work only investigates the impact of data representations on the accuracy of \glspl{DNN} but does not examine other relevant metrics such as computational effort, robustness, or interpretability. Altogether, one should investigate further data representations, architectures, and use cases with regard to various metrics.

\section{Conclusion}\label{sec:Conclusion}

The present paper analyzes the impact of data representations on the performance of \Glspl{DNN} at the example of energy time series forecasting.
Based on an overview of exemplary data representations, we select four different data representations, namely the \textit{naive}, the \textit{naive differences}, the \textit{reshaped}, and the \textit{reshaped differences} data representation. We evaluate these data representations using two different \Gls{DNN} architectures and three forecasting horizons on real-world energy time series.

The results show that, depending on the forecast horizon, the same data representations can have a positive or negative impact on the accuracy of \Glspl{DNN} in the considered energy forecasting use case. Overall, there is no best performing data representation for all forecasting horizons. For example, reshaping the energy time series decreases the forecasting accuracy up to 27\% compared to the \textit{naive} data representation. However, reshaping the differences of the energy time series is beneficial for one-hour ahead forecasts and yields an up to 8\% higher forecasting accuracy compared to the \textit{naive} data representation.

In future work, we plan to investigate the impact of various data representations on \Glspl{DNN} for datasets from different domains and other \Gls{DNN} architectures. In the case of energy forecasting, for example, data representations for additional information like weather could be investigated. For datasets from the given and other domains, the impact of data representations could also be evaluated regarding the problem complexity and dataset size as well as other metrics such as computational effort and interpretability. Furthermore, one could verify the results reported in the present paper using different \Gls{DNN} architectures and datasets. Lastly, it could be interesting to also investigate data representations within \Glspl{DNN} and probabilistic data representations.

\section*{Acknowledgements}

This project is funded by the Helmholtz Association’s Initiative and Networking Fund through Helmholtz AI, the Helmholtz Association under the Program “Energy System Design”, and the German Research Foundation (DFG) under Germany’s Excellence Strategy – EXC number 2064/1 – Project number 390727645.




\end{document}